\documentclass{article}
\usepackage{spconf,amsmath,graphicx}
\usepackage{tikz}
\usepackage{amssymb, bm}
\usepackage{booktabs}

\newcommand{\tikzcircle}[2][black,fill={rgb:orange,2;yellow,2}]{\tikz[baseline=-0.5ex]\draw[#1,radius=#2] (0,0) circle ;}%
\newcommand{\tikzsquare}[2][blue,fill=blue]{\begin{tikzpicture}
\node[rectangle,
    draw = {rgb:white,2;green,2;blue,4},
    text = olive,
    fill = {rgb:white,2;green,2;blue,4}] (r) at (0,0) {};
\end{tikzpicture}}%

\newcommand{\tikzcircletext}[2][blue,fill=blue]{\begin{tikzpicture}
\node[circle,
    draw = {rgb:black,1},
    text = white,
    fill = {rgb:black,1},] (c) at (0,0) {1};
\end{tikzpicture}}%

\title{Deformable VisTR: Spatio temporal deformable attention for video instance segmentation}
\name{Sudhir Yarram{$^\dag$}, Jialian Wu{$^\dag$}, Pan Ji{$^\ddag$}, Yi Xu{$^\ddag$},  Junsong Yuan{$^\dag$}}
  
  \address{$^{\dag}$ University at Buffalo  \qquad 
      $^{\ddag}$OPPO US Research Center, InnoPeak Technology Inc.}
\begin{document}
%
\maketitle
\begin{abstract}
Video instance segmentation (VIS) task requires classifying, segmenting, and tracking object instances over all frames in a video clip. Recently, VisTR \cite{vistr} has been proposed as end-to-end transformer-based VIS framework, while demonstrating state-of-the-art performance. However, VisTR is slow to converge during training, requiring around 1000 GPU hours due to the high computational cost of its transformer attention module. To improve the training efficiency, we propose Deformable VisTR, leveraging spatio-temporal deformable attention module that only attends to a small fixed set of key spatio-temporal sampling points around a reference point. This enables Deformable VisTR to achieve linear computation in the size of spatio-temporal feature maps. Moreover, it can achieve on par performance as the original VisTR with 10$\times$ less GPU training hours. We validate the effectiveness of our method on the Youtube-VIS benchmark. Code is available at {\color{magenta}{https://github.com/skrya/DefVIS}}.
\end{abstract}

\vspace{-5pt}
\section{Introduction}
\label{sec:intro}

Video instance segmentation task \cite{vis_paper} requires classifying, segmenting instances in each frame, and tracking the same instance across frames. It is more challenging as we need to perform instance segmentation for each individual frame and also establish data association of instances across consecutive frames \emph{a.k.a}., tracking. Different from previous methods that rely on sophisticated pipelines \cite{maskprop, vis_paper, stemseg}, the recent work VisTR \cite{vistr} was proposed as an end-end trainable framework to achieve state-of-the-art results. Despite the interesting design and good performance, VisTR requires much longer training time (epochs) to converge. For example, on Youtube-VIS benchmark \cite{vis_paper}, VisTR needs around 1000 GPU hours (500 training epochs\footnote{The epoch number is 18 in VisTR codebase but actually equals to $\sim$500 epochs in common practice, as it trains each video $\sim$28 times in one epoch.}) to converge on a NVIDIA Tesla V100 GPU. This issue mainly arises due to the shortcoming of transformer attention modules to process spatio-temporal clip level features. During initialization, attention weights are uniformly distributed to all the pixels in the feature maps and long training hours is necessary for attention weights to be learned to focus on specific pixels. Also, the computational complexity of attention weights computation in Transformer encoder is of $O(H^{2}W^{2}T^{2}C)$, where  $H, W, T, C$ corresponds to the height, weight, temporal span, and channel dimension of the features belonging to the input clip, respectively. Therefore, it requires  quadratic computation with respect to pixel numbers in the input clip (Table~\ref{tab:withvistr}). Thus, it is of very high computational complexity to process a video clip. 

In the image domain, deformable convolution \cite{deformableconv} can be seen as self-attention mechanism that can attend to sparse meaningful locations. Leveraging it, Deformable DETR \cite{deformabledetr} mitigates the slow convergence and high complexity of DETR \cite{detr}, a transformer framework for object detection. Since, VisTR is inspired from DETR, it is natural that deformable attention can also be extended to address the issues of VisTR. However, Defomable DETR was proposed for image domain, it is not directly applicable to video domain which needs to address both spatial and temporal dimension.

\begin{table*}[htbp!]
  \centering
  \vspace{-5pt}
  \begin{tabular}{l | c | c | c | c }
    \toprule
    Method & Comp. Complexity & Training time (GPU Hours) & Training Epochs & Accuracy (mAP(\%)) \\
    \midrule
    VisTR \cite{vistr} & $O(H^{2}W^{2}T^{2}C)$ & 1000  & $\sim$500 & 35.6 \\ 
   
    \midrule
    \bf{Deformable VisTR} & $O(HWTCK)$ & \bf{120} & \bf{50} & 34.6 \\
    \bottomrule
  \end{tabular}
  \vspace{-6pt}
  \caption{\textbf{Training convergence comparision of VisTR and Deformable VisTR using ResNet-50 backbone on Youtube-VIS val set.} $H, W, T, C$ corresponds to the feature map height, weight, temporal span and channel dimension for the input video clip, respectively. Each reference point attends to a set of $K$ locations out of all spatio-temporal ($HWT$) locations. GPU hours are evaluated on a NVIDIA Tesla V100 GPU. Our proposed Deformable VisTR can achieve on par performance with 10$\times$ less training epochs and time in GPU hours. }
  \vspace{-10pt}
  \label{tab:withvistr}
\end{table*}

 In this paper, we propose Deformable VisTR, which alleviates the high complexity and slow convergence issues of VisTR. It combines the spatio-temporal sparse sampling of deformable convolution and relation modeling capability of transformers to achieve linear computation with respect to pixel numbers. This brings significant improvement from quadratic complexity of VisTR.
 
 To implement Deformable VisTR, we propose a spatio-temporal deformable attention module, which attends to small set of key sampling locations (say $K$, where $K << HWT$) out of all the spatio-temporal feature map pixels ($HWT$) for each reference point. Moreover, these prominent key sampling points are predicted from the feature at the reference point. By replacing the attention modules used in the transformer encoder and decoder of VisTR with spatio-temporal deformable attention module, the computation complexity can be reduced to $O(HWTCK)$, thus achieving linear computation in spatio-temporal feature map size of an input clip (see Table~\ref{tab:withvistr}).

We evaluate the effectiveness of our method on Youtube-VIS benchmark. Compared with VisTR, Deformable VisTR can achieve on par performance with 10$\times$ less GPU training hours. 

\vspace{-5pt}
\section{Related Work}
\label{sec:relatedwork}
\vspace{-6pt}
\subsection{Video Instance Segmentation} 
\vspace{-6pt}The VIS task \cite{vis_paper} has received lot of attention recently. Several state-of-the-art methods \cite{vis_paper, maskprop, stemseg} typically develop sophisticated pipelines to tackle it. Top-down approaches \cite{maskprop, vis_paper} adopt the tracking-by-detection paradigm, depending mainly on image-level instance segmentation models \cite{maskrcnn, hybridIS} and complex instance association rules designed by human. Bottom-up approaches \cite{stemseg} differentiate object instances by clustering learned pixel embeddings. These methods need to employ multiple iterations to generate the masks due to heavy reliance on the dense prediction quality, which makes them slow. In contrast, VisTR proposes to build a simple and end-to-end trainable VIS framework. However, VisTR suffers from long training time which we address by leveraging the idea of deformable convolution. We try to build an end-to-end VIS framework, which is efficient and converges fast. For additional implementation details of VisTR, please refer to the manuscript \cite{vistr}.
\vspace{-10pt}
\subsection{Deformable Attention Mechanism}
\vspace{-6pt} As discussed in \cite{zhu2019empirical}, deformable convolution \cite{deformableconv}, a variant of convolution can be seen as self-attention mechanism. Particularly, deformable convolution is shown to operate more efficiently on image recognition than transformer self-attention. Deformable DETR \cite{deformabledetr} proposes deformable attention where each query element focuses on a fixed set of key sampling points predicted from the features of query element. However, deformable attention was proposed for image domain. We extend it to video domain by proposing spatio-temporal deformable attention module. 

\begin{figure*}[thbp!]
\begin{center}
  \includegraphics[width=1.0\linewidth]{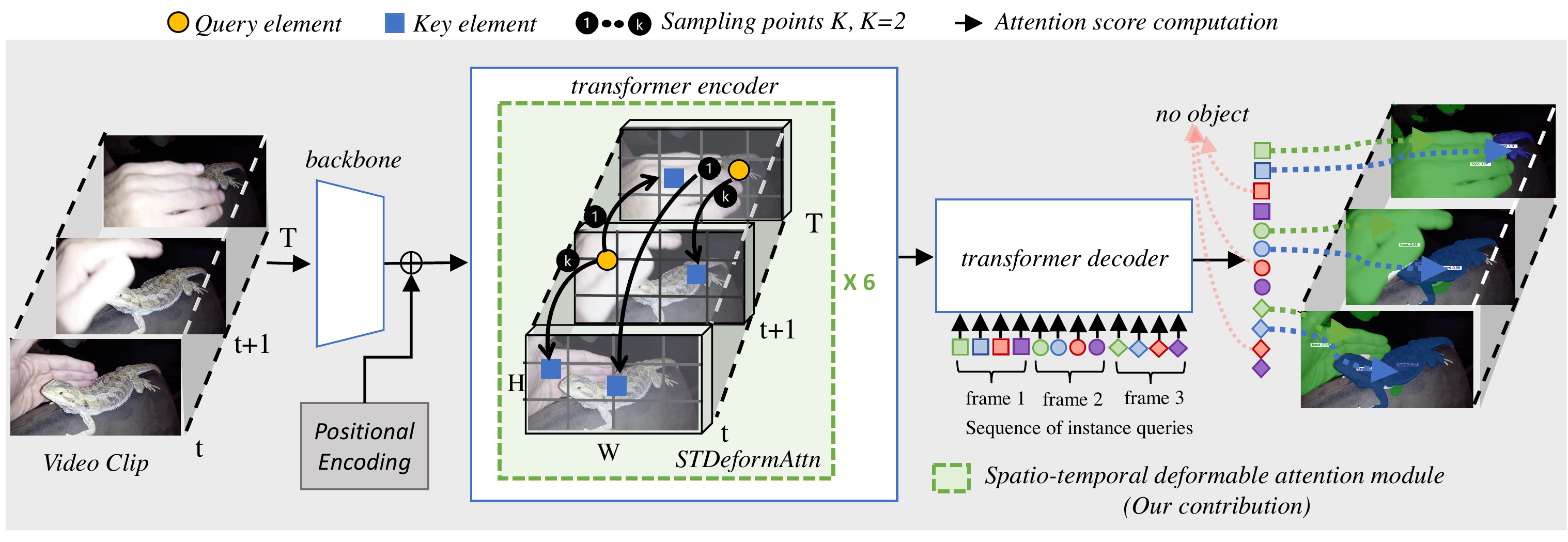}
\end{center}
\vspace{-22pt}
  \caption{\textbf{Illustration of proposed Deformable VisTR.} The green dashed box demonstrates our contribution, spatio-temporal deformable attention (STDeformAttn) module. In transformer encoder of VisTR \cite{vistr}, for each of the query element (\tikzcircle{3pt}), all possible features in spatio-temporal feature map (${HWT}$) are used as key elements in attention score computation. However, in deformable VisTR, for each query element (\tikzcircle{3pt}), only a small fixed set of sampling points ($K, K << HWT$) predicted from query feature are used as key elements (\tikzsquare{}) during attention. In this illustration, $K=3$. Though we do not illustrate, the STDeformAttn module is also used in the transformer decoder. }
\label{fig:how_fig}
\vspace{-10pt}
\end{figure*}
\vspace{-5pt}

\vspace{-10pt}
\section{Proposed Method}
\label{sec:method}

\vspace{-7pt}
\subsection{Revisiting Multi-Head Attention in Transformer}
\vspace{-5pt}
Attention mechanisms is the core component of Transformer network architecture \cite{transformer}. Given a query element (\emph{e.g.,} a target feature corresponding to a 2d pixel position in input image) and a set of key elements (\emph{e.g.,} features belonging to other pixel positions in the same input image), the multi-head-attention module adaptively aggregates the key contents according to the attention weights that measure the agreement of query-key pairs. Different attention heads are used for the model to focus on the contents from different representation subspaces and different  positions. Let $\Omega_{q}$ specify the set of query elements and $q \in \Omega_{q}$ indexes a query element with representation feature $z_{q} \in \mathbb{R}^{C}$, where $C$ is the feature dimension. Let $\Omega_{k}$ specify the set of key elements and $k \in \Omega_{k}$ indexes a key element with representation feature $x_{k} \in \mathbb{R}^{C}$. To differentiate spatial positions, the representation features $z_{q}$ and $x_{k}$ are usually of summation of the element contents and positional embeddings. Then the multi-head attention can be formulated as follows 
 \begin{equation}
    \resizebox{0.9\hsize}{!}{
    $MultiHeadAttn(\bm{z}_{q}, \bm{x}) =  \sum\limits_{m=1}^{M} \bm{W}_{m}\big[ \sum\limits_{k\in \Omega_{k}}A_{mqk} \cdot  \bm{W}_{m}^{'}\bm{x}_{k} \big],$}
    \label{first_eq1}
\end{equation} 
where $m$ indexes the attention head and $M$ is the total number of attention heads. ${\bm{W}_{m}}^{'} \in \mathbb{R}^{C_{v}\times{C}}$ and $\bm{W}_{m} \in \mathbb{R}^{C \times C_{v}}$ are learnable weights. $C_{v} = C/M$. The attention weights  are normalized as $\sum_{k \in \Omega_{k}} A_{mqk} = 1$, where $A_{mqk}  \propto \text{exp} \{{\frac{ {\bm{z}_{q}^{T}} {\bm{U}^{T}_{m}} {\bm{V}_{m} } {\bm{x}_{k}} }{ \sqrt{{C_{v} }}  }} \}$ and $\bm{U}_{m}, \bm{V}_{m} \in \mathbb{R}^{C_{v} \times C}$ are learnable weights.

 While multi head attention modules have shown to be effective, they still need long training schedules before convergence. Suppose the number of query and key elements are of $N_{q}$ and $N_{k}$, respectively. At initialization, $\bm{U}_{m}\bm{z}_{q}$ and $\bm{V}_{m}\bm{x}_{k}$ follow normal distribution with mean of 0 and variance of 1, which makes attention weights $A_{mqk} \approx \frac{1}{N_{k}}$, when $N_{K}$ is large. During backpropogation, it will lead to uncertain gradient computation for input features. Thus, long training hours is necessary for attention weights to be learned to focus on specific keys. In the video domain, where the key elements are usually clip-level spatio-temporal pixels, $N_{k}$ can be very large and convergence is slow. 

Moreover, the computational and memory complexities for the multi-head attention can be very high. For Eq~\ref{first_eq1}, the computational complexity is $O(N_{q}C^{2} + N_{k}C^{2} + N_{q}N_{k}C)$. In the video domain, where the query and key elements are both of video pixels, $N_{q} = N_{k} >> C$, the complexity is dominated by the third term, as $O(N_{q}N_{k}C)$. 

VisTR exploits a standard Transformer encoder-decoder architecture to process input video clip whose features obtained from backbone $\in \mathbb{R}^{C \times T \times H \times W}$, where  $H$, $W$, $T$ denote the feature map height, width, and temporal-span respectively. For the transformer encoder, $N_{q} = N_{k} = HWT$,  So, the computational complexity is $O(H^{2}W^{2}T^{2}C)$. For the transformer decoder, $N_q = N$, where $N$ is the number of object queries (\emph{e.g.,} 360) and $N_{k} = HWT$. The computational complexity is $O(NHWTC)$. Thus, the multi-head attention module suffers from a quadratic complexity growth with the clip-level feature map size. For this reason, VisTR needs long training schedules (approx. 500 epochs and 1000 GPU hours) to achieve optimal performance. 

 The main issue of applying transformer multi-head attention is that it would consider all possible spatial and temporal locations in attention computation. Inspired by Deformable DETR \cite{deformabledetr}, which applies deformable attention to attend to small sample of key sampling points around a reference point, we extend the idea to spatio-temporal deformable attention that is necessary to tackle video instance segmentation. The spatio-temporal deformable attention module only attends to a small number of fixed key sampling points around a query (reference) point, covering both the spatial and temporal directions. By decreasing the number of key elements for each query element from $HWT$ to a small fixed number of keys $K$, we can mitigate the issues of slow convergence.
 
 \subsection{Spatio-Temporal Deformable Attention}
 A crucial component to achieving video instance segmentation is to effectively model both spatial context and temporal context. Incorporating it, a query element is a 3-d reference point that can sample a fixed set of key points in a spatio-temporal set of $H*W*T$ points.
 
 Let $q$ index a query element at a 3-d reference point $p_{q}$, where $p_{q}$ indexes the horizontal, vertical and temporal positional information. Let  $z_{q}$ represent the context feature of $q$. Then, the spatio-temporal deformable attention (STDeformAttn) feature is computed as follows 
 \begin{equation}
    \vspace{-2pt}
    \begin{split}
    STDeformAttn(\bm{z}_{q}, \bm{p}_{q}, \bm{x}) =\\  \sum\limits_{m=1}^{M} \bm{W}_{m}[ \sum_{k=1}^{K}A_{mqk} \cdot \bm{W}_{m}^{'}\bm{x}(\bm{p}_{q} + \Delta \bm{p}_{mqk}) ].
    \end{split}
    \label{first_eq2}
\end{equation}

where $k$ indexes the sampled keys and $K$ is the total sampled key number ($K << HWT$) and $m$ indexes the attention head. The scalar weight $A_{mqk}$ denote the attention weight of the $k^{th}$ sampling point in the $m^{th}$ attention head, where $A_{mqk}$ is normalized by $\Sigma_{k=1}^{K} A_{mqk} = 1$. $A_{mqk}$ are predicted via linear projection over the query feature $\bm{z}_{q}$. $\Delta \bm{p}_{mqk}$ denote the sampling offset of the $k^{th}$ sampling point in the $m^{th}$ attention head. $\Delta \bm{p}_{mqk} \in \mathbb{R}^{3}$ are of 3-d real numbers with unconstrained range. As $\bm{p}_{q} + \Delta \bm{p}_{mqk}$ is fractional, bilinear interpolation is applied as in computing $\bm{x}(\bm{p}_{q} + \Delta \bm{p}_{mqk})$. $\Delta \bm{p}_{mqk}$ is obtained via linear projection over the query feature $\bm{z}_{q}$.

 The computational complexity of the STDeformAttn module is of $O(2N_{q}C^{2} +  N_{q}MKC)$, where $N_{q}$ is the number of query elements. Replacing the attention module in VisTR encoder with STDeformAttn module, where $N_{q} = HWT$ and $MK > C$, the complexity becomes $O(HWTCK)$. So, the computational complexity drops from $O(H^{2}W^{2}T^{2}C)$ to linear complexity in the spatial-temporal feature size. When it is applied to VisTR decoder the complexity becomes $O(NKC)$, where $N_{q} = N $ ($N$ is the number of object queries). The proposed STDeformAttn module becomes equivalent to VisTR Transformer attention module if $K = HWT$.

\vspace{-7pt}
\subsection{Deformable Transformer Encoder}
\vspace{-5pt}
We replace the Transformer attention modules in the VisTR encoder and decoder with the proposed STDeformAttn module (see Fig.~\ref{fig:how_fig}). Both the input and output of the encoder are same resolution spatio-temporal feature maps.  In encoder, we extract features using ResNet \cite{resnet} backbone. These features are transformed to contain $C$ channels by a $1 \times 1$ convolution. In application of the STDeformAttn module in encoder, the output are spatio-temporal features with same resolutions as the input. Both the query and key elements are of pixels from the spatio-temporal feature maps. For each query pixel, the reference point is itself. To identify which spatio-temporal position the feature corresponds to, precise position information is needed. To incorporate this, we use fixed positional encoding information that contains the three dimensional (horizontal, vertical and temporal) positional information in the clip,\ to supplement the features.
\vspace{-7pt}
\subsection{Deformable Transformer Decoder}
\vspace{-5pt} Transformer decoder consists of cross-attention and self-attention modules. In cross-attention modules, the object queries are the query elements, while feature maps from the encoder are used as key elements. In the self-attention modules, object queries are both query and key elements. Since our proposed STDeformAttn module is primarily designed for processing convolutional feature maps, we only replace each cross-attention module to be the STDeformAttn module.

By replacing the Transformer attention modules with STDeformAttn module in transformer of VisTR, we establish an efficient and fast converging video instance segmentation system, termed as Deformable VisTR (see Fig.~\ref{fig:how_fig}).

\vspace{-5pt}
\section{Experiments and Results}
\label{sec:experiments}
\vspace{-7pt}
\subsection{Datasets}
\vspace{-5pt}
 YouTubeVIS dataset \cite{vis_paper} is first dataset for video instance segmentation which contains 2238 training, 302 validation, and 343 test video clips. It contains 40 object categories along with per frame instance masks. The evaluation metrics are average precision (AP) with the video intersection over Union (IoU) of the mask sequences as the threshold. \vspace{-7pt}\subsection{Implementation Details}\vspace{-5pt} We use ImageNet \cite{imagenet} pre-trained ResNet-50 \cite{resnet} without FPN \cite{fpn} as the backbone for ablations. $M = 8$, $K=32$, $C=384$ are set for STDeformAttn module by default and is initialized randomly. Unless mentioned otherwise, we mainly follow VisTR \cite{vistr} in training strategy, loss functions, and hyper parameter setting. The models are trained for 50 epochs and we decay learning rate by 0.1 at the 40th epoch. Following VisTR, we use Adam optimizer with backbones's learning rate as $1 \times 10^{-5}$, base learning rate of $1 \times 10^{-4}$, $\beta_{1} = 0.9$, $\beta _{2} = 0.999$, and the weight decay of $10^{-4}$. We multiply the learning rates for the linear modules used for predicting object query sampling offsets by 0.1. Model is trained on 4 NVIDIA Tesla V100 GPUs of 16GB each and run time is evaluated on a single GPU. \vspace{-7pt}
\subsection{Comparison with VisTR}
\vspace{-5pt}
As shown in Table~\ref{tab:withvistr}, Deformable VisTR achieves on par performance as VisTR with $10 \times$ less training GPU hours and epochs. This shows that our proposed STDeformAttn module can mitigate the issue of convergence successfully.

   
\vspace{-7pt}
\subsection{Comparison with State of the Art}
\vspace{-5pt}
Table~\ref{tab:sota} provides comparison with state-of-the-art methods. We do not compare with \cite{maskprop, Reduce-propose} due to their heavy component design, which results in low inference ($<$ 6 FPS) speed. In Table~\ref{tab:sota}, VisTR and Deformable VisTR are end-to-end trainable frameworks, while other methods are not. Our method achieves 34.6\% mAP. Overall, our method is a fully end-to-end framework that converges 10$\times$ times faster than VisTR. 
\begin{table}[thbp!]
  \centering
   \vspace{-18pt}
  \begin{tabular}{l | c | c | c | c }
    \toprule
    Method & \begin{tabular}{@{}c@{}}{Fully} \\ {End-to-End}\end{tabular}    & Aug. & FPS & AP  \\
    \midrule
    $\text{MaskTrack \cite{vis_paper}}_{_{\textit{ CVPR'19}}}$ & & & 28.6  & 30.3   \\
    $\text{SipMask \cite{cao2020sipmask}}_{_{\textit{ ECCV'20}}}$ &  &  \checkmark & 34.1 & 33.7    \\
    $\text{STEmSeg \cite{stemseg}}_{_{\textit{ ECCV'20}}}$ &  & \checkmark  \checkmark & 4.4 & 30.6    \\
    $\text{CompFeat~\cite{compfeat}}_{_{\textit{ AAAI'21}}}$ &  & \checkmark  \checkmark & 32.8 & 35.3\\
    $\text{SGNet~\cite{sgnet}}_{_{\textit{ CVPR'21}}}$ &  & \checkmark \checkmark  & {19.8} & 34.8     \\
    $\text{STMask~\cite{stmask}}_{_{\textit{ CVPR'21}}}$ &  &  & {28.6} & 33.5     \\
    $\text{CrossVIS~\cite{crossvis}}_{_{\textit{ ICCV'21}}}$ &  &  & \bf{39.8} & 34.8     \\
    $\text{QueryInst \cite{Fang_2021_ICCV}}_{_{\textit{ ICCV'21}}}$ &  &   & 32.3 & 34.6  \\
   \midrule
    $\text{VisTR \cite{vistr}}_{_{\textit{ CVPR'21}}}$  & \color{red}{\checkmark} &  \checkmark  & 30.0  & \bf{35.6}   \\ 
    
    \midrule
    \bf{Deformable VisTR} & \color{red}{\checkmark} &  \checkmark & 33.0 & 34.6 \\
    \bottomrule
  \end{tabular}
  \vspace{-6pt}
  \caption{\textbf{Comparison of Deformable VisTR with the state-of-the-art methods on Youtube VIS val set.} All the entries use ResNet-50 \cite{resnet} as backbone. The methods are listed in temporal order. ``\checkmark" indicates multi-scale input images during training. ``\checkmark \checkmark" indicates stronger data augmentation (\emph{e.g.,} additional data \cite{compfeat, stemseg}, random crop\cite{maskprop})}
  \vspace{-10pt}
  \label{tab:sota}
\end{table}
\vspace{-8pt}
\subsection{Ablation Study on STDeformAttn module}
\vspace{-5pt}
Table~\ref{tab:ablation}, presents ablations of varying the number of sampling points $K$ for each reference point in the proposed STDeformAttn module. Increasing the number of sampling points $K$ from 16 to 32 can further improve accuracy by $0.8$ mAP.

\begin{table}[thbp!]
  \centering
  \vspace{-5pt}
  \begin{tabular}{c | c | c }
    \toprule
     backbone & $K$ & AP \\
    \midrule
     ResNet-50 & 16 & 33.8  \\ 
    \midrule
     ResNet-50 & 32 & 34.6   \\
    \bottomrule
  \end{tabular}
  \vspace{-6pt}
  \caption{\textbf{Ablation of STDeformAttn module.} $K$ is the number of key points for each query feature. $K=32$ gives the best result.}
  \vspace{-10pt}
  \label{tab:ablation}
\end{table}
\vspace{-8pt}
\section{Conclusion}
\label{sec:conclusion}
\vspace{-10pt}
Deformable VisTR is designed to be an efficient, fast-converging and end-end trainable video instance segmentation framework. To implement Deformable VisTR, we propose spatio-temporal deformable attention modules, which is an efficient attention mechanism in processing clip-level feature maps. We achieve comparable performance as VisTR with 10$\times$ less GPU training hours.\\

\noindent \textbf{Acknowledgement.} This work is supported in part by a gift grant from OPPO and National Science Foundation Grant CNS1951952.

\bibliographystyle{IEEEbib}
\bibliography{ReviewTemplate}

\end{document}